\documentclass{article}
\usepackage{spconf,amsmath,graphicx}
\usepackage{booktabs}
\belowdisplayskip \abovedisplayskip

\newlength{\sectionReduceTop}
\newlength{\sectionReduceBot}
\newlength{\subsectionReduceTop}
\newlength{\subsectionReduceBot}
\newlength{\abstractReduceTop}
\newlength{\abstractReduceBot}
\newlength{\captionReduceTop}
\newlength{\captionReduceBot}
\newlength{\subsubsectionReduceTop}
\newlength{\subsubsectionReduceBot}

\newlength{\eqnReduceTop}
\newlength{\eqnReduceBot}

\newlength{\horSkip}
\newlength{\verSkip}

\newlength{\figureHeight}
\usepackage{graphicx} 
\usepackage{subcaption}
\usepackage{multicol}
\usepackage{booktabs} 
\usepackage{algorithm} 
\usepackage{algorithmic} 
\usepackage{color}
\usepackage{amsmath}
\usepackage{amssymb}
\usepackage{amsfonts}
\usepackage{multirow, varwidth}
\usepackage{stfloats} 
\usepackage{verbatim}
\makeatletter
\newif\if@restonecol
\makeatother
\usepackage{xr}
\usepackage[amsmath,thmmarks]{ntheorem}

\usepackage[dvipsnames]{xcolor}
\usepackage{enumitem}
\usepackage{flushend}

\usepackage{times}
\usepackage{latexsym}
\usepackage[T1]{fontenc}
\usepackage[utf8]{inputenc}
\usepackage{microtype}
\usepackage{inconsolata}

\usepackage{url}

\usepackage{xspace}
\makeatletter
\DeclareRobustCommand\onedot{\futurelet\@let@token\@onedot}
\def\@onedot{\ifx\@let@token.\else.\null\fi\xspace}

\def\eg{\emph{e.g}\onedot} 
\def\ie{\emph{i.e}\onedot}

\def\etc{\emph{etc}\onedot}

\makeatother

\newcommand{\todo}[1]{}  

\newcommand{\emptydraft}[1]{{\color{black}{}}}

\newcommand{\abr}[1]{\textsc{#1}}

\newcommand{\reffig}[1]{Fig.~\ref{#1}}

\usepackage{booktabs,arydshln}
\makeatletter
\def\adl@drawiv#1#2#3{%
        \hskip.5\tabcolsep
        \xleaders#3{#2.5\@tempdimb #1{1}#2.5\@tempdimb}%
                #2\z@ plus1fil minus1fil\relax
        \hskip.5\tabcolsep}
\newcommand{\cdashlinelr}[1]{%
  \noalign{\vskip\aboverulesep
           \global\let\@dashdrawstore\adl@draw
           \global\let\adl@draw\adl@drawiv}
  \cdashline{#1}
  \noalign{\global\let\adl@draw\@dashdrawstore
           \vskip\belowrulesep}}
\makeatother

\usepackage[misc]{ifsym} 

\newcommand{\mn}{\abr{IMU2CLIP}\xspace} 
\newcommand{\clip}{\abr{CLIP}\xspace}

\title{\mn: Multimodal Contrastive Learning for IMU Motion Sensors from Egocentric Videos and Text Narrations}

\name{
    \begin{tabular}{@{}c@{}}
        Seungwhan Moon$^{*}$, Andrea Madotto$^{*}$,
        Zhaojiang Lin, Alireza Dirafzoon, Aparajita Saraf\\
        Amy Bearman, Babak Damavandi
    \end{tabular}
    \thanks{*: Joint First Authors.}
}
\address{
    Meta Reality Labs \\
}

\begin{document}

\maketitle

\begin{abstract}



We present \mn{}, a novel pre-training approach to align Inertial Measurement Unit (IMU) motion sensor recordings with video and text, by projecting them into the joint representation space of Contrastive Language-Image Pre-training (\clip).
The proposed approach allows \mn{} to translate human motions (as measured by IMU sensors) into their corresponding textual descriptions and videos -- while preserving the \textit{transitivity} across these modalities.

We explore several \textit{new} IMU-based applications that \mn{} enables, such as motion-based media retrieval and natural language reasoning tasks with motion data.
In addition, we show that \mn{} can significantly improve the downstream performance when fine-tuned for each application (\eg activity recognition), demonstrating the universal usage of \mn{} as a new pre-trained resource.
Our code will be made publicly available.







\end{abstract}


\section{Introduction}
\label{sec:introduction}
\vspace{0pt}

With the growing popularity of smart glasses or new-generation wearable devices, \textit{first-person} or \textit{egocentric} videos have recently become much more prevalent than ever before \cite{ego4d, epickitchen21, aria_pilot_dataset}.
These egocentric videos are often accompanied by the parallel head-mounted IMU sensor readings, which record devices' linear and rotational movements and accelerations. 

Given its low power consumption and low privacy implications, IMU is regarded as an important modality for powering various on-device models that require understanding of device wearer's movement patterns (\eg exercise / activity recognition for health applications).
The previous works on IMU modeling typically focus on the purpose-built datasets with manual annotations \cite{jiang2022multi, chen2021probability}, which are limited in their scale. 
Consequently, the utilization of IMU models in real-world scenarios has been confined to a relatively small number of use cases. 

On the contrary, for the modalities that are widely studied (\eg text, video), there are vast large-scale resources such as BERT \cite{bert} and GPT \cite{GPT} for text, or CLIP4Clip \cite{CLIP4Clip} for videos.
These powerful pre-trained resources have driven the development of many application-oriented models, showing significant improvements when fine-tuned for each respective task \cite{dodge2020fine}. 
To the best of our knowledge, however, the study on the equivalent resources for encoding IMU signals has been lacking.

\begin{figure}[t]
    \centering
    \includegraphics[width=\columnwidth]{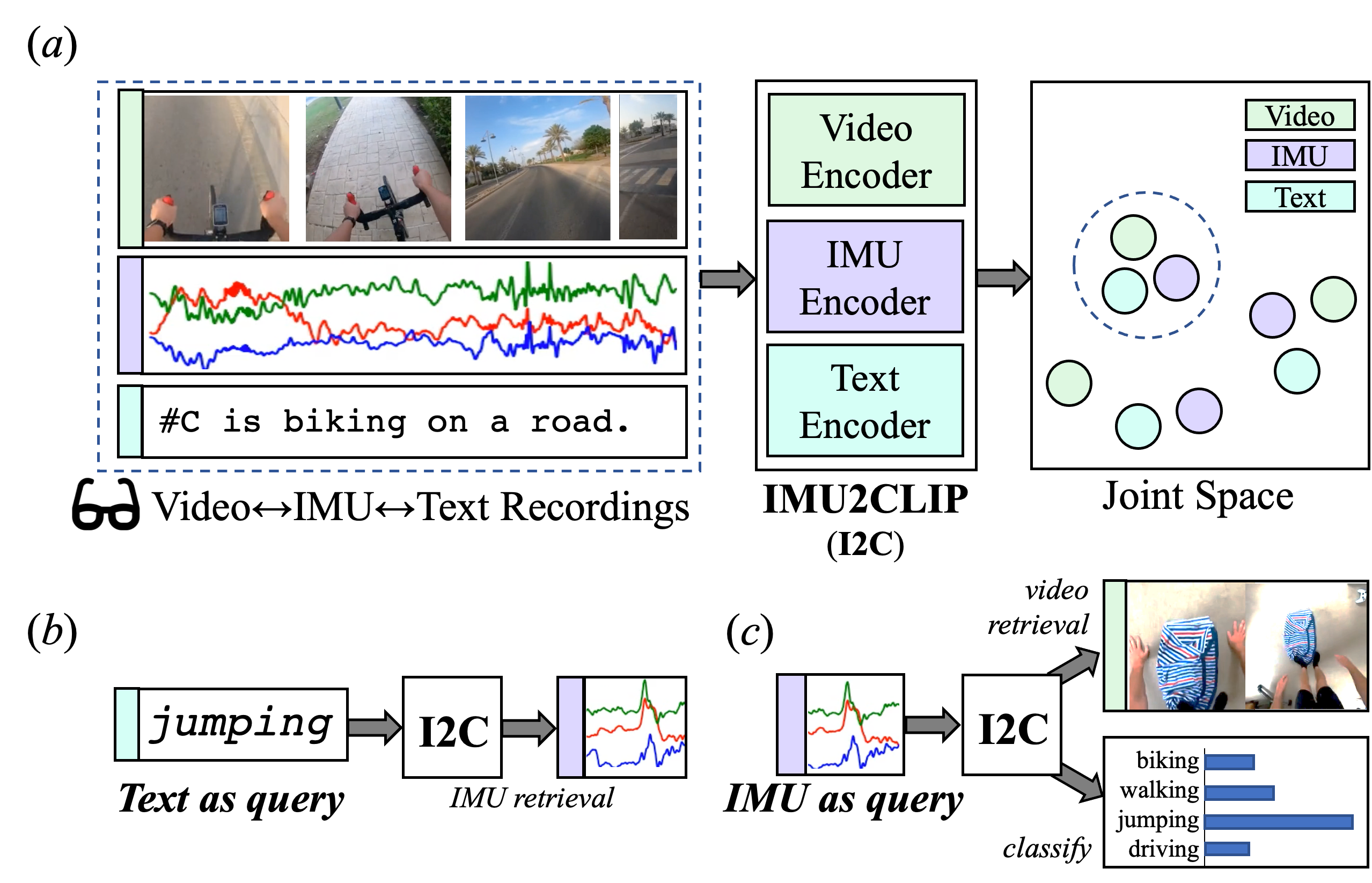}
    \vspace{-4pt}
    \caption{Illustration of \mn (I2C): (\textit{a}) The model aligns the parallel video$\leftrightarrow$IMU$\leftrightarrow$text data in the joint space. Once trained, \mn is used as a retriever for both (\textit{b}) IMU and (\textit{c}) videos, or as a classifier for downstream applications.}
    \vspace{-16pt}
    \label{fig:teaser}
\end{figure}

Inspired by the recent works that leverage large pre-trained models for other modalities, we present \mn, a new approach to pre-train an IMU encoder by aligning the parallel Video $\leftrightarrow$ IMU $\leftrightarrow$ Text data in an un-supervised manner via multimodal contrastive training.
Specifically, we propose to use \clip \cite{clip}, which contains the video encoder and the language model pre-trained on the large parallel image-text data, from which the IMU encoder can learn a semantic representation of various scenes transferred from other modalities. 

To show the efficacy of the proposed approach, we evaluate our models on several benchmark tasks as well as new applications that \mn enables, such as IMU-based media retrieval, leveraging the modality-transitivity that \mn exhibits (\reffig{fig:teaser}).
Most importantly, we show that the fine-tuned \mn can significantly improve the performance of several downstream tasks, when compared to the identical IMU model trained from scratch.

\textbf{Our contributions} are as follows: (1) We propose a novel large-scale pre-training approach for IMU sensors, and release the resulting large pre-trained IMU encoders for future research. (2) We provide an in-depth empirical analysis evaluating the pre-trained models, for both upstream and downstream fine-tuning tasks. (3) Lastly, we present novel applications that show the feasibility of a wider usage for IMU sensor signals.

\vspace{0pt}
\section{Related Work}
\label{sec:related_work}
\vspace{0pt}

\noindent{\textbf{Contrastive Learning}}
    is as an efficient self-supervised framework applied across multiple domains, which learns similar/dissimilar representations from data that are organized into similar/dissimilar pairs. 
    For instance, SimCLR \cite{simclr} is a unimodal application of contrastive learning in the data augmentation setting, in which the authors propose to learn a vision encoder given a set of perturbed images.
    As an example in multimodal settings, Contrastive Language–Image Pre-training (CLIP) \cite{clip, lin2022egocentric}  learns visual representations from natural language supervision using image and text pairs, achieving competitive results in \eg zero-shot image classification, image retrieval via text, and image/caption generation. 
    Similarly, WAV2CLIP \cite{wav2clip} proposes to learn audio representation by distilling it from CLIP. 
    We extend this line of work on contrastive learning to a unique multimodal setting that utilizes IMU signals, which is specific to a new generation of devices (such as smart glasses) that are equipped with such sensors.

\begin{figure*}[t!]
    \centering
    \includegraphics[width=0.93\textwidth]{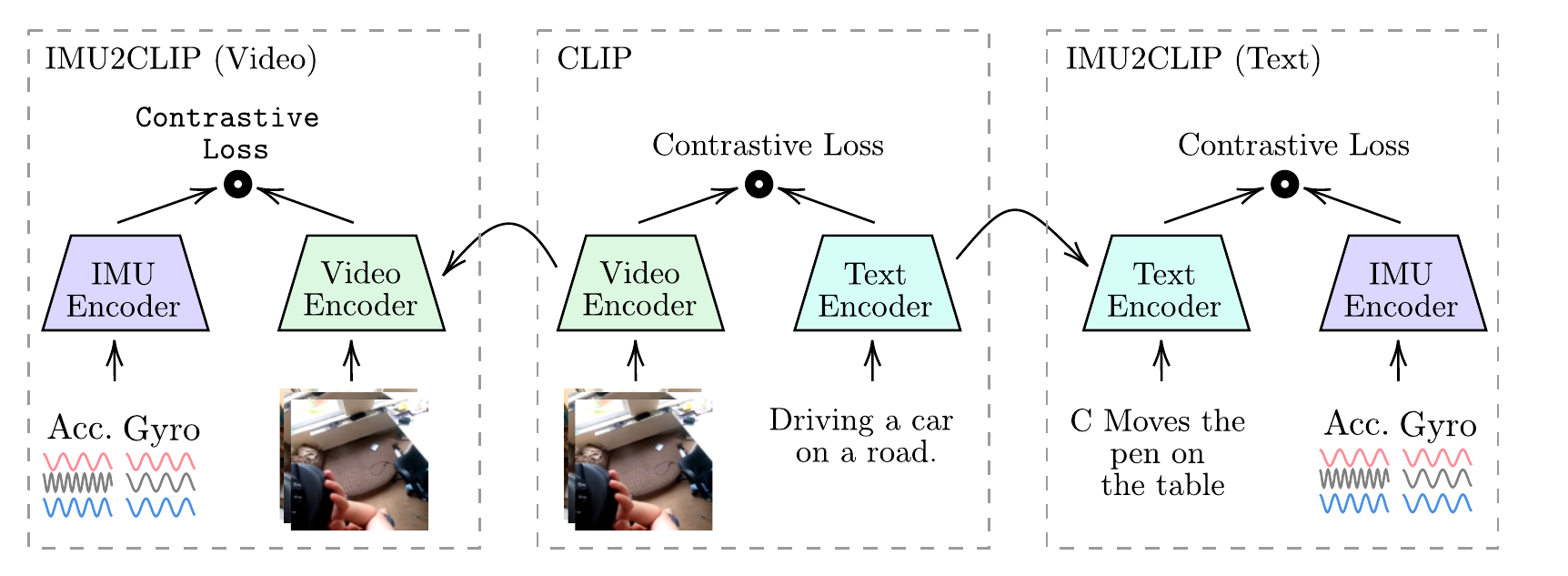}
    \caption{Illustration of the proposed multimodal contrastive learning for \mn.  \clip~\cite{radford2021learning} is used to align IMU$\leftrightarrow$Video (left), and IMU$\leftrightarrow$Text (right). \includegraphics[height=1em]{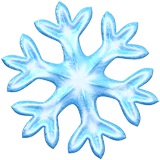}: the parameters of the encoder are frozen during training.
    }
    \label{fig:contrastive_learning}
\end{figure*}

\noindent{\textbf{Pre-training Resources}}:
    There are numerous pre-trained resources for well-studied modalities such as images or text.
    Many popular computer vision models (\eg ResNet \cite{resnet}) are typically trained on large supervised datasets such as ImageNet \cite{imagenet}, \etc.
    For language processing, the most popular language models (LM) include BERT \cite{bert, reimers-2019-sentence-bert}, GPT-2 \cite{GPT2}, and GPT-3\cite{GPT3}, which typically use self-superivsion techniques such as next-word predictions or masked token predictions, thus without any explicit task labels.
    Studies report that these pre-trained resources achieve competitive zero-shot performance \cite{clip}, and when fine-tuned, often outperform fully supervised models on several downstream tasks \cite{dodge2020fine}.
    
    To our knowledge, the equivalent resource for encoding IMU signals is not made publicly available.
    Inspired by this line of work, we propose to perform large-scale pre-training for the unique sensor (IMU) signals dataset, and show that such pre-training significantly improves the performance for the downstream applications as well. 

\noindent{\textbf{Egocentric Datasets}}: 
We are particularly interested in egocentric (first-person) datasets, for understanding of users’ activities from head-mounted devices.
Several data collection efforts have been made for building egocentric datasets, including
Ego4D \cite{ego4d}, Epic-Kitchens \cite{epickitchen21}, and Aria \cite{aria_pilot_dataset} datasets. 

Using these datasets, we propose various sub-tasks that can effectively evaluate diverse capabilities of \mn{}, and demonstrate the feasibility of future applications.
In addition, we implement a universal multimodal data loader to allow for easy cross-modality and cross-domain (dataset) studies.

\noindent{\textbf{IMU Modeling}}: 
IMU signals have been widely used in various motion recognition tasks, such as pose estimation~\cite{zihajehzadeh2017gaussian}, walking speed estimation~\cite{zihajehzadeh2017gaussian}, foot placement prediction~\cite{chen2021probability}. Various deep learning architectures have been explored for modeling IMU in downstream tasks, including Transformer-CNN based IMU models~\cite{jiang2022multi} for gesture recognition, 1D-CNN and GRU ensemble IMU models~\cite{kim2021wearable} for clinical balance assessment, and Bi-LSTM IMU models~\cite{ashry2020charm} for human activity recognition.
Our work proposes a new IMU model architecture, and conducts ablation studies over other models above. 
Different from prior work modeling IMU in a specific task, however, our work focuses on learning general IMU representations by aligning IMU with other modalities (\eg images and text), which can enable wider downstream applications.


\section{Methods}
\label{sec:methods}
\vspace{0pt}

For IMU pre-training, we propose to align the parallel portions of the IMU $\leftrightarrow$ Video $\leftrightarrow$ (optionally) Text data, using multimodal cross-modal contrastive learning schemes \cite{clip, CLIP4Clip}.
In a nutshell, we train an IMU encoder such that the IMU representation of a given clip resembles the representation of its corresponding video frames, and optionally, corresponding textual descriptions or narrations (Section \ref{subsec:method:loss}). 
We perform ablation studies over multiple IMU encoder architectures, as detailed in Section \ref{subsec:method:encoder}.
\reffig{fig:contrastive_learning} illustrates the overall approach.

\subsection{Cross-modal Contrastive Learning for IMU}
\label{subsec:method:loss}

We consider a batch of $B$ ground-truth \underline{I}MU$\leftrightarrow$\underline{V}ideo$\leftrightarrow$\underline{T}ext parallel windows: $\{(\mathbf{i}_1, \mathbf{v}_1, \mathbf{t}_1), ..., (\mathbf{i}_B, \mathbf{v}_B, \mathbf{t}_B)\}$, where the embeddings of each modality lies on the unit hypersphere $S^D$. 
Since the embeddings are unit-normalized, the similarity can be simply calculated as their inner product: 
\begin{align}
    \text{sim}(\mathbf{i}_i, \mathbf{v}_j) &= \langle \mathbf{i}_i,  \mathbf{v}_j \rangle \\
    \text{sim}(\mathbf{i}_i, \mathbf{t}_j) &= \langle \mathbf{i}_i,  \mathbf{t}_j \rangle
\end{align}

We can then define the IMU-to-Video ($\mathbf{i}2\mathbf{v}$) and IMU-to-Text ($\mathbf{i}2\mathbf{t}$) retrieval distributions based on the cross-modal similarities across the parallel signals:
\begin{align}
    P_{\mathbf{i}2\mathbf{v}}(\mathbf{v}_j | \mathbf{i}_i) &= \frac{\exp(\text{sim}(\mathbf{i}_i, \mathbf{v}_j))^{1/\gamma}}{\sum_{k=1}^B \exp(\text{sim}(\mathbf{i}_i, \mathbf{v}_k))^{1/\gamma}} \\
    P_{\mathbf{i}2\mathbf{t}}(\mathbf{t}_j | \mathbf{i}_i) &= \frac{\exp(\text{sim}(\mathbf{i}_i, \mathbf{t}_j))^{1/\gamma}}{\sum_{k=1}^B \exp(\text{sim}(\mathbf{i}_i, \mathbf{t}_k))^{1/\gamma}}
\end{align}
where $\gamma$ is a temperature parameter that controls the concentration of the distributions.

We then train three flavors of \mn: (a) aligning IMU$\leftrightarrow$Video, (b) IMU$\leftrightarrow$Text (when the text narration data are available), and (c) IMU$\leftrightarrow$Video$\leftrightarrow$Text.
Specifically, we propose to project the IMU representations into the joint \clip space \cite{clip} to leverage the visual and textual knowledge already encoded in \clip.
Similar to \cite{CLIP4Clip, clip}, we propose to minimize the symmetric cross-modal contrastive loss.
For the IMU$\leftrightarrow$Video alignment, we use the sum of IMU-to-Video and Video-to-IMU cross-entropy losses:
\begin{align}
    \mathcal{L}_{\mathbf{i}2\mathbf{v}} 
    &= -\frac{1}{B} \sum_{i=1}^B \log \frac{\exp(\text{sim}(\mathbf{i}_i, \mathbf{v}_i))^{1/\gamma}}{\sum_{k=1}^B \exp(\text{sim}(\mathbf{i}_i, \mathbf{v}_k))^{1/\gamma}} \nonumber \\
    \mathcal{L}_{\mathbf{v}2\mathbf{i}} 
    &= -\frac{1}{B} \sum_{i=1}^B \log \frac{\exp(\text{sim}(\mathbf{i}_i, \mathbf{v}_i))^{1/\gamma}}{\sum_{k=1}^B \exp(\text{sim}(\mathbf{i}_k, \mathbf{v}_i))^{1/\gamma}} \nonumber \\
    \mathcal{L}_{\mathbf{i}\leftrightarrow\mathbf{v}} &= \frac{1}{2} (\mathcal{L}_{\mathbf{i}2\mathbf{v}} + \mathcal{L}_{\mathbf{v}2\mathbf{i}}) 
    \label{eq:contrastive_loss}
\end{align}

The loss for IMU$\leftrightarrow$Text alignment ($\mathcal{L}_{\mathbf{i}\leftrightarrow\mathbf{t}}$) can be defined similarly, and consequently $\mathcal{L}_{\mathbf{i}\leftrightarrow\mathbf{v}\leftrightarrow\mathbf{t}}=\mathcal{L}_{\mathbf{i}\leftrightarrow\mathbf{v}}+\mathcal{L}_{\mathbf{i}\leftrightarrow\mathbf{t}}$.
To preserve the text-vision alignment that \clip already exhibits, we freeze the parameters of the image and text \clip encoders.

\noindent \textbf{Implementation details}. To expedite the training, we pre-process each media to have equal-sized parallel windows (IMU $\leftrightarrow$ Video $\leftrightarrow$ Text).
The data module retrieves the parallel data of a requested window size at a given timestamp, and caches them for faster training.
In addition, to accommodate the memory constraints, we pool the negative samples within the same batch (randomly shuffled), reducing the load on each GPU.
We optimize the parameters with Adagrad \cite{Adagrad} with batch size 16, learning rate 0.01, epsilon $10^{-8}$, and decay 0.1.





\subsection{IMU Encoder Architectures}
\label{subsec:method:encoder}

\begin{figure}[t!]
    \centering
    \includegraphics[width=0.7\columnwidth]{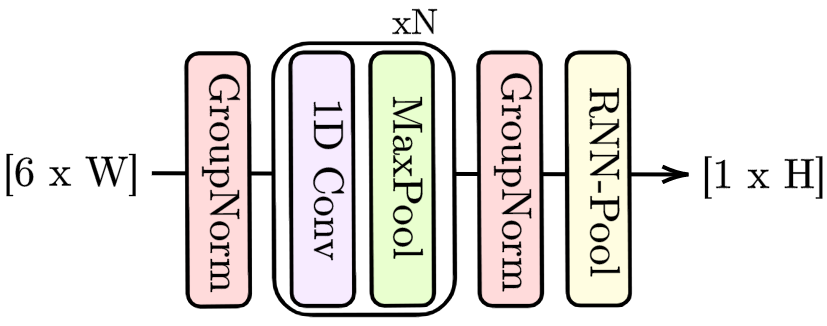}
    \caption{
        Illustration of the Stacked RNN architecture for the IMU encoder used in IMU2CLIP.
    }
    \label{fig:imu_encoder}
\end{figure}

For the IMU encoder, we propose a stack of 1D-CNNs and RNN-based architecture (Figure \ref{fig:imu_encoder}), which performed the best in our ablation studies\footnote{Ablation results can be found:  \url{tinyurl.com/imu2clip-experiments}}. First, we perform a GroupNorm operation  to normalize the Accelerometer (3D) and the Gyroscope (3D) signals independently.
We then perform a stack of $N$ 1D-CNN, a Max Pooling with kernel size 5, and then another GroupNorm to normalize the output features.
Finally, we use an RNN (\ie GRU in our experiments) to combine the CNN output, and thus generating the final embedded representation. 
\todo{include ablation studies results}


\section{Experiments}
\label{sec:experiments}
\vspace{0pt}

\begin{table}[t]
\centering
\scalebox{0.9}{
    \begin{tabular}{@{}lccc@{}}
    \toprule
    \multicolumn{1}{c}{\textbf{Ego4d}}              & \textbf{Tra.} & \textbf{Val.} & \textbf{Tst.} \\
    \midrule
    \# of Media files                      & 1444          & 161           & 688           \\
    Total Media Durations                    & 540h          & 60h           & 265h          \\
    \# of IMU$\leftrightarrow$Text/Video windows (5s) & 528K          & 68K           & 266K          \\
    \# of IMU$\rightarrow$4 classes  windows (5s)   & 1552          & 760           & 241           \\
    \bottomrule
    \toprule
    \multicolumn{1}{c}{\textbf{Aria}}               & \textbf{Tra.} & \textbf{Val.} & \textbf{Tst.} \\
    \midrule
    \# of Media files               & 747           & 259           & 277           \\
    Total Media Durations                    & 138h          & 43h           & 51h           \\
    \# of IMU$\leftrightarrow$Video windows (1s)    & 496K          & 157K          & 184K          \\
    \# of IMU$\rightarrow$5 classes windows (1s)     & 25K           & 138K          & 162K          \\ 
    \bottomrule
    \end{tabular}
}
\vspace*{\captionReduceTop}
\caption{
    Dataset Statistics for Ego4D and Aria.
}
\vspace*{\captionReduceBot}
\vspace{-6pt}   
\label{tab:dataset_statistics}
\end{table}


\subsection{Dataset}

We use Ego4D \cite{ego4d} and Aria \cite{aria_pilot_dataset} as the main datasets for the experiments below,
both of which feature parallel video and IMU signals. 
For Ego4D, a subset of the clips are also annotated with their corresponding narrations.
We split the data into train, validation, and test sets (split by video IDs).  
The statistics of the datasets are provided in Table \ref{tab:dataset_statistics}.

\subsection{Tasks}


Note that the proposed pre-training approach enforces the alignment of the IMU, video, and text representations, which allows for new and unique cross-modal applications.
We propose the following real-world downstream applications as the novel tasks to evaluate the performance of the IMU encoders.


\noindent \textbf{Task 1. IMU Retrieval via Textual Queries (Text$\rightarrow$IMU)}, 
where the goal is to retrieve a window of IMU signals given free-form textual queries.
Once the IMU signals are retrieved, we can also retrieve the corresponding videos, thus allowing for a new and power-efficient way of performing media retrieval or online action detection.
The retrieval performance is measured on the held-out test set (Recall@$k$ and Mean Reciprocal Rank (MRR)), using the text narrations as queries and the IMU signals as the retrieval pool. 

\noindent \textbf{Task 2. Video Retrieval based on IMU  (IMU$\rightarrow$Video)}, 
where the goal is to retrieve videos based on IMU signals, allowing for an intuitive way of analyzing motion signals data.
We measure the performance on the held-out test set, using the IMU signals as queries and the videos as the retrieval target.

\noindent \textbf{Task 3. IMU-based Activity Recognition},
where the goal is to predict a discrete activity label given a window of IMU signals. 
We use the manual motion annotations for Aria (\eg hiking, running, biking) and soft annotations for Ego4D via text matching of the narrations provided.

\subsection{Results}



\begin{table*}[t]
\begin{center}
    \scalebox{0.75}{
    \begin{tabular}{@{}ccc cccc cccc cccc cccc@{}}
    \toprule
    \multicolumn{3}{c}{\textbf{Train Modalities}} & \multicolumn{4}{c}{\textbf{IMU $\rightarrow$ Text}}                               & \multicolumn{4}{c}{\textbf{Text $\rightarrow$ IMU}}  & \multicolumn{4}{c}{\textbf{IMU $\rightarrow$ Video}}                              & \multicolumn{4}{c}{\textbf{Video $\rightarrow$ IMU}}                               \\
    \midrule
    \midrule
    IMU           & Video         & Text         & R@1  & R@10  & R@50 & MRR & R@1   & R@10  & R@50 & MRR & R@1  & R@10  & R@50 & MRR & R@1   & R@10  & R@50 & MRR \\
    \cmidrule(r){1-3}
    \cmidrule(r){4-7}
    \cmidrule(r){8-11}
    \cmidrule(r){12-15}
    \cmidrule(r){16-19}
    \checkmark    & \checkmark    &              & 4.86 & 18.75 & 48.26                        & 0.104                      & 4.17  & 15.62 & 43.06                        & 0.084                & 9.06 & 43.13 & 78.75                        & 0.2011                      & 12.19 & 45.31 & 80.00                           & 0.226       \\ 
    \checkmark    &               & \checkmark   & 5.21 & 25.00    & 60.42                        & 0.123                      & 7.29  & 28.82 & 60.07                        & 0.143                & 3.75 & 25.94 & 62.81                        & 0.105                      & 3.75  & 24.06 & 56.88                        & 0.098      \\
    \checkmark    &    \checkmark    & \checkmark   & 4.52 & 22.91 & 56.60                        & 0.118                      & 5.90  & 22.92 & 56.60                        & 0.139                & 8.75 & 40.63 & 73.44                        & 0.183                      & 11.56  & 42.19 & 75.94                        & 0.213      \\    
    \midrule
    \toprule
    \multicolumn{3}{c}{}                        & \multicolumn{4}{c}{\textbf{(Video $\rightarrow$ Text)}}                              & \multicolumn{4}{c}{\textbf{(Text $\rightarrow$ Video)}} & \multicolumn{8}{c}{\multirow{2}{*}{-}}                 \\
    \cmidrule(r){1-3}
    \cmidrule(r){4-7}
    \cmidrule(r){8-11}
    \small{(\clip)} & \checkmark & \checkmark            & 6.94  & 32.29 & 64.24 & 0.150        & 8.33 & 33.68 & 65.28 & 0.168                       & & & & & & & &                     \\ 
    \bottomrule
    \end{tabular}
    }
\end{center}
    \vspace*{\captionReduceTop}
    \caption{
            Text$\leftrightarrow$IMU and Video$\leftrightarrow$IMU retrieval performances of the pre-trained \mn models on Ego4D, with different modalities used for training. The last row shows the video retrieval performance of OpenAI’s \clip model on the same test set.
    }
\label{tab:exp:retrieval} 
\end{table*}

Table \ref{tab:exp:retrieval} shows the IMU$\leftrightarrow$Text and IMU$\leftrightarrow$Video retrieval performance on the Ego4D test set, of \mn trained via different combinations of modalities.

\begin{figure}[t!]
    \centering
    \includegraphics[width=\columnwidth]{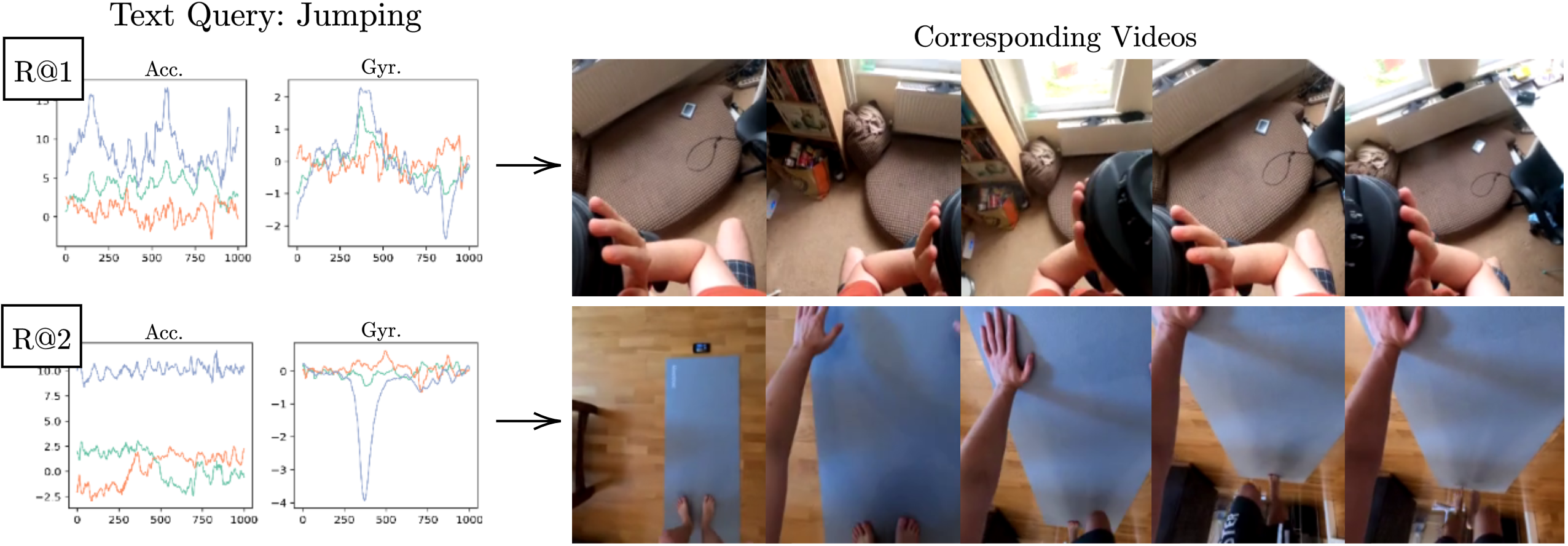}
    \caption{Illustration of IMU-based media retrieval. Given a free-form textual query (\eg ``\textit{jumping}"), (Left): \mn's predictions of the semantically closest IMU signals from the Ego4D test set (top-2). (Right): the gold-parallel videos corresponding to the retrieved IMU signals (as a reference). The retrieved media match the semantics of the input query.
    }
    \label{fig:demo_t2i}
\end{figure}

\noindent \textbf{Results 1: IMU-based media search with textual queries}:
The Text$\rightarrow$IMU column in Table \ref{tab:exp:retrieval} shows performances on Task 1.
Note that the \clip embeddings already exhibit the Video $\leftrightarrow$ Text transitivity, and thus  \mn trained using IMU $\leftrightarrow$ Video achieves a competitive zeroshot performance for IMU $\leftrightarrow$ Text retrieval as well.
When text narrations are used for pre-training, the model achieves an even higher recall performance. See \reffig{fig:demo_t2i} for visualizations\footnote{For better readability, we also provide the animated GIF visualizations for all experiments at: \url{tinyurl.com/imu2clip-visualizations}}.




To help contextualize the recall performances in Table \ref{tab:exp:retrieval}, we also show (as a reference) the Text $\leftrightarrow$ Video retrieval performance of the near-SOTA video encoder (\clip) (bottom).
The narrow margin in the performances (\eg MRR=0.143 for Text→IMU \textit{vs}. MRR=0.168 for Text→Video) shows that the IMU encoder could serve as a power-efficient alternative for a video encoder in many applications.

\begin{figure}[t!]
    \centering
    \includegraphics[width=\columnwidth]{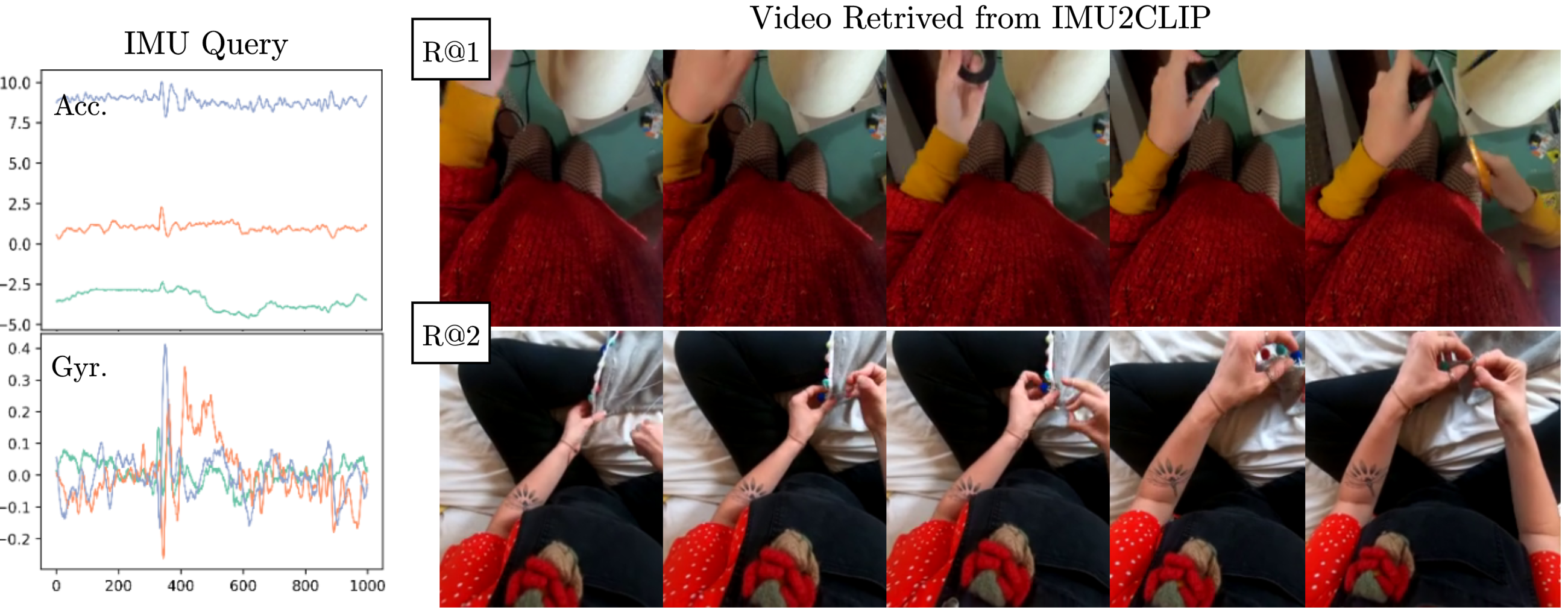}
    \caption{Illustration of the IMU to Video Retrieval. (Top): IMU signals and their corresponding ground-truth video. (Bottom): \mn’s model predictions of their corresponding videos from the Ego4D test set (top-5), given the IMU signals. It can be seen that the videos retrieved based on IMU are visually and semantically similar to the gold video.
    }
    \label{fig:demo_i2v}
\end{figure}

\noindent \textbf{Results 2: We can search for videos, given IMU recordings}.
The IMU$\rightarrow$Video column in Table \ref{tab:exp:retrieval} shows the Ego4D performances on Task 2 (See \reffig{fig:demo_i2v} for examples).
We observe higher recall performances in general, showing that the IMU signals and videos have a higher compatibility.
When the model is trained on all three modalities, we observe competitive results across all tasks, while the best performances are from the bi-modal models aligned with each respective task.

We observe similar patterns on the Aria data as well:
for the IMU$\rightarrow$Video retrieval, \mn achieves MRR=0.182, and R@$\{1,10,50\}$ of $\{8.48, 38.83, 77.67\}$, respectively.
For the Video$\rightarrow$IMU retrieval, \mn achieves MRR=0.190, and R@$\{1,10,50\}$ of $\{8.48, 44.19, 78.57\}$. 
Note that the Aria dataset does not have text narrations annotated. 

\begin{table}[t]
    \centering
    \scalebox{0.93}{
        \begin{tabular}{@{}rrcccc@{}}
        \toprule
        & \multirow{2}{*}{\textbf{Models}} &  \multicolumn{2}{c}{\textbf{Ego4D}} & \multicolumn{2}{c}{\textbf{Aria}} \\ 
            \cmidrule(r){3-4}
            \cmidrule(r){5-6}
          & & F1    & Acc. & F1 & Acc. \\  
        \midrule
        \multicolumn{2}{r}{Random Init. IMU Encoder}    & 23.23 & 49.92 & 56.35 & 76.11  \\
        \midrule
        \multirow{3}{1.85cm}{\mn ($\mathbf{i}\leftrightarrow\mathbf{v}$)} & + Zeroshot                  & 19.39 & 23.08 & 18.46 & 21.52  \\
        & + Probing                   & 40.55 & 61.46 & \textbf{62.52} & \textbf{83.54}  \\
        & + Fine tuning               & 43.07 & \textbf{65.87} & 61.77 & 82.31 \\
        \midrule
        \multirow{3}{1.85cm}{\mn ($\mathbf{i}\leftrightarrow\mathbf{t}$)} & + Zeroshot                  & 31.89 & 36.38 & - & -  \\
        & + Probing                   & 45.12 & 58.01 & - & -  \\
        & + Fine tuning               & \textbf{45.15} & 63.14 & - & -  \\ 
        \midrule
        \multirow{3}{1.85cm}{\mn ($\mathbf{i}\leftrightarrow\mathbf{v}\leftrightarrow\mathbf{t}$)} & + Zeroshot  & 27.24 & 24.26 & - & -  \\
        & + Probing                   & 42.16 & 55.21 & - & -  \\
        & + Fine tuning               & 44.17 & 62.66 & - & -  \\         
        \bottomrule
        \end{tabular}
    }
    \vspace*{\captionReduceTop}
     \caption{
            IMU-based activity recognition on Ego4D and Aria datasets, comparing the randomly initialized model and the pre-trained \mn models, with IMU$\leftrightarrow$Video, IMU$\leftrightarrow$Text and IMU$\leftrightarrow$Video$\leftrightarrow$Text pre-training. \textbf{Bold} denotes the best performance for each metric: F1 and Accuracy (Acc). 
        }
    \vspace*{\captionReduceBot}
    \label{tab:exp:finetuning}    
\end{table}

\noindent \textbf{Results 3: Fine-tuned \mn significantly outperforms the vanilla model with the same architecture, on downstream tasks.}
Table \ref{tab:exp:finetuning} shows the activity recognition results on Ego4D and Aria datasets.
For all experiments, we use the same IMU architecture (Stacked RNN).
For zeroshot experiments, we encode the surface names of each activity (\eg hiking) with the \clip text encoder, and use the nearest-neighbor classifier on the projected IMU embeddings (thus not using any supervision labels).
Probing adds a linear layer on top of the IMU encoder while keeping the IMU encoder frozen, and for fine-tuning we allow all parameters of the encoder to be trainable.
The consistent improvements in the fine-tuning performances (\eg $\sim$16 points absolute improvement in accuracy for Ego4D, comparing the randomly initialized vanilla model \textit{vs.} fine-tuned model) show that \mn can learn high quality representations for IMU signals. 

Comparing the pre-trained models trained via various combinations of modalities again shows that \mn preserves the transitivity among modalities (video $\leftrightarrow$ text $\leftrightarrow$ IMU).

\begin{figure}[t!]
    \centering
    \includegraphics[width=\columnwidth]{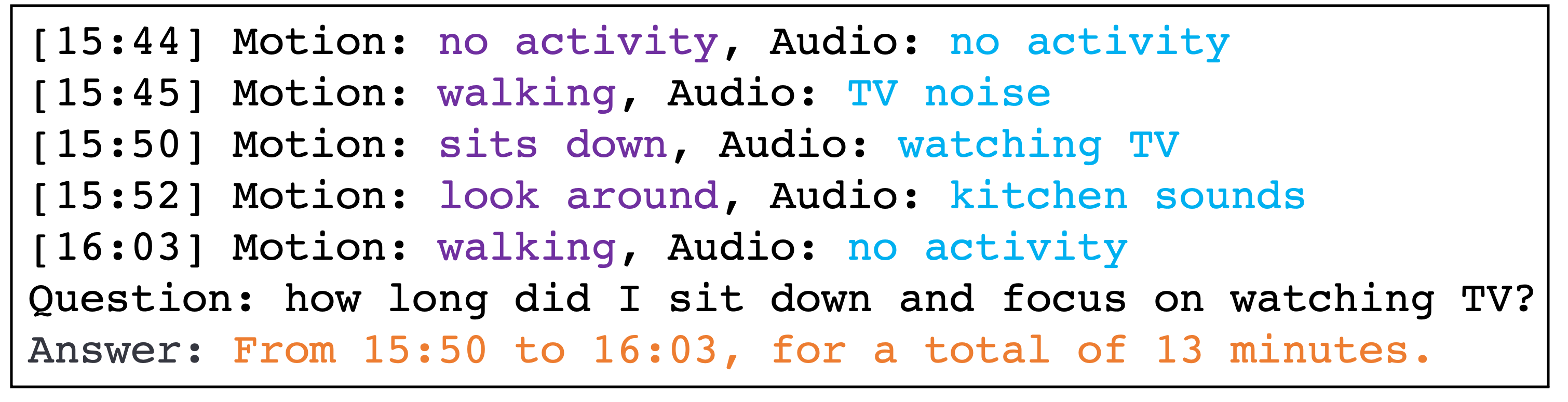}
    \caption{
        Demonstration of an LM-based multimodal reasoning model, using two ambient sensors: \textcolor{Purple}{IMU} and \textcolor{Cyan}{audio}.
        Given the sensor logs and the question, LM generates a \textcolor{orange}{response} grounded on the multimodal context.
    }
    \label{fig:socratic_demo}
\end{figure}

\subsection{Qualitative Analysis: Multimodal Reasoning with Ambient Sensors}

Further exploring the benefit of \mn that translates sensor signals into text, we demonstrate a multimodal reasoning model that operates only on the ambient sensor logs (\reffig{fig:socratic_demo}).
Specifically, we run \mn as a zeroshot tagging model on the clips from Ego4D, to obtain textual descriptions of the IMU and the Audio sensor readings.
We then use a large LM (\ie GPT-3 \cite{GPT3}) to use the sensor logs as conditioning context to answer summarizing questions such as: ``\textit{What can you tell me about the user activity using the motion logs?}", or memory recall prompts such as: ``\textit{What time did I start biking?}"
The LM then generates a response via a causal language inference step, which completes the process for zeroshot multimodal reasoning.
Unlike the similar approach such as Socratic Models \cite{socratic}, the proposed approach does not rely on the video signals at all -- which would incur much higher power consumption -- thus operating better under real-world constraints.


\section{Conclusions}
\label{sec:conclusions}

With the growing popularity of wearable devices of diverse form factors (\eg smart glasses), it is important to study the capability of the ambient sensors such as IMU motion signals. 
To this end, we propose a new multimodal contrastive training approach for representing IMU signals, and release the pre-trained IMU encoders to be used for future research.
Our empirical analysis highlights the efficacy of the proposed approach on many existing and new IMU-based applications.
In addition, we show that \mn can significantly improve the downstream performance when fine-tuned, demonstrating the universal usage of \mn.

\bibliographystyle{IEEEbib}
\bibliography{bibliography}

\end{document}